# Adaptive Cost-Sensitive Machine Learning for Autonomous Robot Navigation Failure Prediction: When Not All Errors Are Equal

Rifa Ferzana[1]

[1] School of Informatics, University of Edinburgh, Edinburgh, UK

**Abstract.** Autonomous robot navigation failures differ not only in categorical severity but also in the physical context in which they occur. A near-miss at low speed under reliable sensing is not equivalent to the same event during rapid motion, close obstacle approach or degraded perception. This paper reframes navigation failure prediction as consequence-sensitive forecasting. We first establish a fixed baseline in which training weights are modulated by categorical severity, then introduce an adaptive extension defining a state-dependent consequence function combining severity with normalised velocity, obstacle proximity and sensing uncertainty, together with a risk-sensitivity term that rises as conditions deteriorate. We evaluate on 2,000 simulated differential-drive episodes (~1,000,000 timesteps) using episode-level GroupKFold, with external validation on the UCI SCITOS G5 dataset. Fixed weighting raises Logistic Regression high-severity recall from 0.851 to 0.985 and reduces missed consequence cost from 1,940 to 313; the adaptive extension reaches 0.998 and 82. Under matched false-positive conditions, however, the discriminative advantage is modest (0.986 versus 0.984), so most of the gain reflects a more conservative operating point rather than better ranking. The effect is consistent across all five folds and stable across a threefold span of context coefficients. Because the primary simulation produced no collisions, we add a controlled extension in which 108 of 600 episodes terminate in contact: collision recall rises from 0.850 to 0.966 (fixed) and 0.984 (adaptive), with missed collision cost falling from 1,000 to 105, at false-positive rates of 0.413 and 0.799, respectively. Context-dependent consequence modelling thus provides a principled mechanism for allocating conservatism by physical risk.

## 1 Introduction

Autonomous robot navigation has advanced considerably in recent years, driven by improvements in sensor technology, simultaneous localisation and mapping (SLAM) and learning-based control [1, 2]. However, the question of how to predict and prevent navigation failures remains underexplored relative to the effort devoted to improving navigation performance itself.

A critical limitation of existing approaches is their reliance on symmetric evaluation metrics: accuracy, precision and F1 treat all errors as equally costly, yet a slight path deviation incurs a minor efficiency cost whilst an undetected imminent collision risks damaging the robot, its environment or nearby humans. This asymmetry has been recognised in other safety-critical domains; in healthcare AI, decision-theoretic frameworks encoding differential misclassification costs substantially improve the allocation-relevance of predictive models [3].

The traditional framing of robot navigation failure prediction asks a binary question: will the robot fail? We argue that this is insufficient. A more useful question for safety-critical deployment is: will the robot fail and how serious would failing to recognise that event be? This reframing, from binary failure classification to consequence-sensitive forecasting, motivates the present work.

In this paper, we propose a cost-sensitive classification framework in which training sample weights are modulated by the consequence cost of each failure type. Rather than optimising for aggregate accuracy, our approach explicitly upweights high-severity events during training, encouraging classifiers to prioritise the detection of dangerous failures at a modest cost to overall performance.

Our contributions are as follows. First, we formulate robot navigation failure prediction as consequence-sensitive prediction rather than conventional binary failure classification. Second, we define three safety-oriented evaluation metrics, high-severity recall (HSR), cost detection rate (CDR) and cost-weighted F1 (CW-F1), that capture safety-relevant performance characteristics obscured by standard metrics. Third, we present an empirical evaluation using both simulated robot navigation trajectories (2,000 episodes, 1,000,000 timesteps) and the UCI Wall-Following Robot dataset (SCITOS G5, 24 ultrasound sensors), demonstrating that cost-sensitive training achieves substantial improvements in high-severity failure detection across multiple classifier architectures. Fourth, we provide sensitivity analyses for the risk-sensitivity parameter, cost-ratio configuration and context coefficients, together with paired fold-level and risk-stratified analyses that establish where and why context adaptation helps.

## 2 Related Work

### 2.1 Robot Navigation and Failure Detection

Classical navigation relies on geometric path planning [4], potential fields [5] and probabilistic localisation [6]; learning-based navigation [7, 8] introduces new failure modes. Sünderhauf et al. [9] examined introspective capabilities for autonomous robots and Daftry et al. [10] learned to predict failures in vision-based navigation, but both frame detection as binary classification without distinguishing severity.

### 2.2 Safe Reinforcement Learning

The safe RL community has developed constrained optimisation frameworks [11, 12] and benchmarks such as Safety-Gymnasium [13] and OmniSafe [14]. These learn safe policies rather than predicting failures in deployed systems; our work is complementary, providing a supervisory layer operating alongside any navigation stack.

### 2.3 Cost-Sensitive Learning

Cost-sensitive classification has a well-established literature [15, 16]. Elkan [17] provided foundational analysis; subsequent work explored cost-sensitive decision trees [18], boosting [19] and deep networks [20]. In robotics, these approaches remain underutilised, with most failure prediction systems evaluated using metrics that ignore differential consequences. Recent healthcare AI work [3] demonstrated the value of encoding consequence asymmetry into training; we extend this to autonomous robotics, where failures carry physical consequences.

# 3 Methodology

## 3.1 Problem Formulation

We frame navigation failure prediction as a binary classification problem with asymmetric misclassification costs. Let $x(t)$ denote the feature vector extracted from the observation window $[t-w, t]$ and let $y(t) \in \{0, 1\}$ indicate whether any failure occurs in the interval $[t, t+h]$. Associated with each positive instance is a consequence cost $c(t)$ reflecting the severity of the worst failure in the lookahead window.

We define three severity levels based on the robot's state relative to obstacles. A path deviation (cost = 1) is triggered when the robot strays more than 3 metres from the straight-line trajectory connecting the start and goal positions; this represents an efficiency failure that does not endanger the robot or its environment. A near-miss (cost = 5) is triggered when the robot approaches within 0.8 metres of an obstacle surface; this represents a safety-relevant event that, whilst not resulting in contact, indicates degraded navigation performance. A collision (cost = 25) is triggered when the robot contacts an obstacle at a distance below 0.2 metres; this represents a critical safety failure with potential for physical damage. The cost ratio of 1:5:25 reflects a reasonable engineering judgement; we evaluate sensitivity to alternative ratios in Section 5.5.

## 3.2 Cost-Sensitive Sample Weighting

Standard supervised classification treats all training samples equally or, in the case of class-balanced training, adjusts weights solely to account for class imbalance. Our approach goes further by modulating sample weights according to the consequence cost of each failure. For a training sample i with label $y(i)$ and consequence cost $c(i)$, we define the sample weight as: $w(i) = 1 + \lambda \cdot c(i)$, where $\lambda$ is a risk-sensitivity hyperparameter controlling the degree to which high-consequence failures are upweighted during training. When $\lambda = 0$, all samples receive equal weight and training reduces to standard classification. As $\lambda$ increases, the classifier is increasingly penalised for misclassifying high-cost failures, at the potential expense of performance on lower-cost events. We normalise the weights such that they have unit mean across the training set.

We set $\lambda = 2.0$ throughout our primary experiments; this value was fixed a priori so that a collision sample (cost = 25) receives approximately 50 times the weight of a normal sample and was not tuned to optimise any reported metric. We evaluate sensitivity to $\lambda$ in Section 5.5.

This formulation is model-agnostic: any classifier that accepts sample weights during training can be adapted to the cost-sensitive framework. For models that do not natively support sample weighting (such as scikit-learn's MLP implementation), we employ cost-proportional oversampling as an approximation. We note that oversampling changes the training distribution rather than preserving it, which introduces a confounding factor: observed improvements in the MLP may partially reflect the effects of oversampling itself. Production implementations should prefer per-sample loss scaling (e.g. via PyTorch's weighted cross-entropy) to isolate the cost-sensitivity effect.

### 3.3 Context-Dependent Consequence

Categorical severity alone does not capture the full risk of a navigation event. A near-miss at low speed with reliable sensing differs materially from the same categorical event during rapid motion or degraded perception. We therefore define a state-dependent consequence function scaling base severity by physical context: $C_{dyn} = C_{base} [1 + \alpha V + \beta U + \gamma P]$, where V is normalised speed (current velocity divided by maximum speed, clipped to [0,1]), U is sensing uncertainty (standard deviation of range readings across sensors, normalised) and $P = 1 - \min(d, d_{max})/d_{max}$ is proximity risk, with d the minimum range reading and $d_{max}$ = 5 m. Context terms are scaled to [0,1] using training-fold statistics only. We fix $\alpha = 0.5$, $\beta = 1.0$ and $\gamma = 1.5$ a priori, giving greatest emphasis to proximity.

### 3.4 Adaptive Risk Sensitivity

Beyond scaling consequence, we allow the risk-sensitivity parameter to respond to operating conditions: $\lambda_i = \lambda_0 [1 + \eta U + \rho P]$, so risk sensitivity increases with sensing degradation and obstacle proximity. We set $\eta = 1.0$, $\rho = 2.0$ and $\lambda_0 = 1.5$. The adaptive weight is $w_A = 1 + \lambda_i C_{dyn}$, normalised to unit mean within each training fold. This increases conservatism in high-risk states without imposing explicit constraints on the navigation controller.

### 3.5 Simulation Environment

We developed a 2D robot navigation simulator in which a differential-drive robot navigates from a fixed start position (1, 1) to a goal position (19, 19) in a 20×20 metre arena populated with 8 randomly placed circular obstacles (radii 0.3–1.2m, positions sampled uniformly whilst avoiding start and goal regions). The robot is controlled by a potential field navigation algorithm that combines an attractive force toward the goal with repulsive forces from nearby obstacles. The simulator models five range sensors (analogous to LiDAR) arranged at −90°, −45°, 0°, +45° and +90° relative to the robot heading, each with a maximum range of 5 metres and configurable Gaussian noise. The robot operates at a maximum speed of 1.5 m/s with a simulation timestep of 0.1 seconds. Episodes terminate upon reaching the goal (distance < 0.5m) or after 500 timesteps.

To generate diverse failure conditions, we simulate episodes under six configurations: (i) normal operation with low sensor noise ($\sigma = 0.05$, 40% of episodes), (ii) normal operation with elevated sensor noise ($\sigma = 0.15$, 10%), (iii) controller drift with gradual perturbation to navigation commands (15%), (iv) aggressive navigation with reduced obstacle avoidance gain (15%), (v) intermittent sensor failure with stochastic dropout at 30% per-timestep probability (10%) and (vi) combined drift with elevated noise (10%). We generated 2,000 episodes totalling approximately 1,000,000 timesteps. At the timestep level, the failure distribution was: 95.05% normal, 2.89% path deviation, 2.06% near-miss and 0.00% collision. No collision events were recorded: the potential-field controller-maintained clearance above the 0.2 m contact threshold in every episode. The collision category (cost 25) is retained in the framework for platforms with less conservative controllers, but all high-severity samples in our simulated data are near-misses. At the sample level, the 294,000 windows comprise 92.15% normal, 3.94% path deviation and 3.91% near-miss samples.

### 3.6 Feature Extraction

From each window of w = 10 timesteps we extract 21 features in four categories: motion (mean, standard deviation and current linear speed and acceleration; mean, standard deviation and maximum absolute angular velocity), proximity (windowed mean, minimum, current and trend of minimum range; mean range variance; front sensor reading), trajectory (path deviation and trend, heading error, distance to goal, goal progress) and environmental context (sensor noise level).

Range-derived features include additive Gaussian noise ($\sigma$ = 0.1–0.15) preventing ground-truth leakage. The prediction horizon h = 50 timesteps (~5 s) provides a realistic early-warning interval; a stride of 3 reduces redundancy between consecutive samples.

## 4 Experimental Setup

### 4.1 Classifiers

We evaluate Logistic Regression (L2, 1,000 iterations), Random Forest (300 trees, depth 12), XGBoost (300 trees, depth 6, learning rate 0.1) and an MLP (128-64-32, ReLU, early stopping with 15% holdout). Hyperparameters were set to common defaults and held constant across all experiments.

Each classifier is evaluated in two configurations. In the standard configuration, class imbalance is addressed using class-balanced weighting (for LR and RF) or positive-class scaling (for XGB), with no consideration of failure severity. In the cost-sensitive configuration, sample weights are computed using the cost-weighted scheme described in Section 3.2 with $\lambda$ = 2.0.

### 4.2 Evaluation Metrics

In addition to standard metrics (accuracy, precision, recall, F1 and AUROC), we report three safety-oriented evaluation metrics. High-Severity Recall (HSR) is the recall computed exclusively on samples where the consequence cost is 5 or greater (near-miss and collision events); this measures the classifier's ability to detect the failures that matter most for safety. Cost Detection Rate (CDR) is defined as 1 − (total cost of missed failures / total cost of all actual failures); this measures the proportion of total consequence cost that the classifier successfully intercepts. Cost-Weighted F1 (CW-F1) is a modified F1 score in which true positives and false negatives are weighted by $(1 + c(i))$; this provides a single summary statistic that balances precision against cost-weighted recall.

### 4.3 Evaluation Protocol

All experiments use five-fold grouped cross-validation at the episode level (GroupKFold), ensuring that all samples originating from a given navigation episode are assigned exclusively to either the training or validation partition within each fold. Features are standardised to zero mean and unit variance using statistics computed exclusively from the training partition of each fold. The data pipeline proceeds as follows: 1,000,000 simulation timesteps are processed through sliding windows with stride 3 to produce 294,000 candidate prediction samples; majority-class subsampling then yields 53,088 evaluation samples (30,000 normal, 23,088 failure). All reported results represent means ± standard deviations across the five folds.

### 4.4 Adaptive-Method Ablation Protocol

To isolate each component's contribution, we evaluate five matched conditions under identical GroupKFold splits: (i) class-balanced weighting, (ii) threshold-moving on a standard classifier, (iii) fixed consequence weighting ($\lambda = 2.0$), (iv) dynamic consequence only and (v) dynamic consequence with adaptive risk sensitivity. All context parameters remain fixed a priori. Because methods operating at different false-positive rates are not directly comparable, we additionally report a matched-FPR comparison. To avoid calibrating on evaluation data, this uses a nested split: training episodes are divided into fit (80%) and calibration (20%) partitions by episode; models are trained on the fit partition, the target FPR and decision threshold are selected on the calibration partition, the threshold is then frozen and applied once to the held-out test fold.

## 5 Results

### 5.1 Standard Classification Performance

**Table 1.** Standard classification metrics (episode-level GroupKFold CV, 2,000 episodes). Mean ± std. dev.

| Model | Approach | F1 | AUROC | Recall | Prec. |
|---|---|---|---|---|---|
| LR | Std | 0.853±.002 | 0.943±.003 | 0.881±.005 | 0.827±.004 |
| LR | CS | 0.804±.005 | 0.934±.004 | 0.971±.003 | 0.690±.006 |
| RF | Std | 0.984±.001 | 0.999±.000 | 0.995±.002 | 0.973±.002 |
| RF | CS | 0.974±.002 | 0.997±.001 | 0.985±.003 | 0.964±.003 |
| XGB | Std | 0.991±.001 | 0.999±.000 | 0.999±.001 | 0.982±.002 |
| XGB | CS | 0.987±.001 | 0.999±.000 | 0.999±.001 | 0.974±.002 |
| MLP | Std | 0.982±.002 | 0.998±.001 | 0.992±.003 | 0.971±.003 |
| MLP | CS | 0.984±.002 | 0.998±.001 | 0.998±.001 | 0.973±.002 |

XGBoost achieves the highest F1 (0.991) and AUROC (0.999), followed by RF (0.984) and MLP (0.982).

### 5.2 Cost-Sensitive Performance

**Table 2.** Cost-sensitive metrics (episode-level GroupKFold CV). HSR = High-Severity Recall, CDR = Cost Detection Rate, CW-F1 = Cost-Weighted F1.

| Model | Approach | HSR | CDR | CW-F1 | Missed Cost |
|---|---|---|---|---|---|
| LR | Std | 0.851 | 0.860 | 0.905 | 1,940 |
| LR | CS | 0.985 | 0.977 | 0.935 | 313 |
| RF | Std | 0.994 | 0.995 | 0.994 | 72 |
| RF | CS | 0.999 | 0.999 | 0.993 | 10 |
| XGB | Std | 0.998 | 0.998 | 0.996 | 24 |
| XGB | CS | 1.000 | 1.000 | 0.996 | 7 |
| MLP | Std | 0.991 | 0.991 | 0.992 | 124 |

| MLP | CS | 0.997 | 0.996 | 0.994 | 56 |
|---|---|---|---|---|---|

Cost-sensitive LR improves HSR from 0.851 to 0.985 (+13.4 pp). Missed cost drops from 1,940 to 313—a six-fold reduction. MLP improves HSR from 0.991 to 0.997 with no F1 loss.

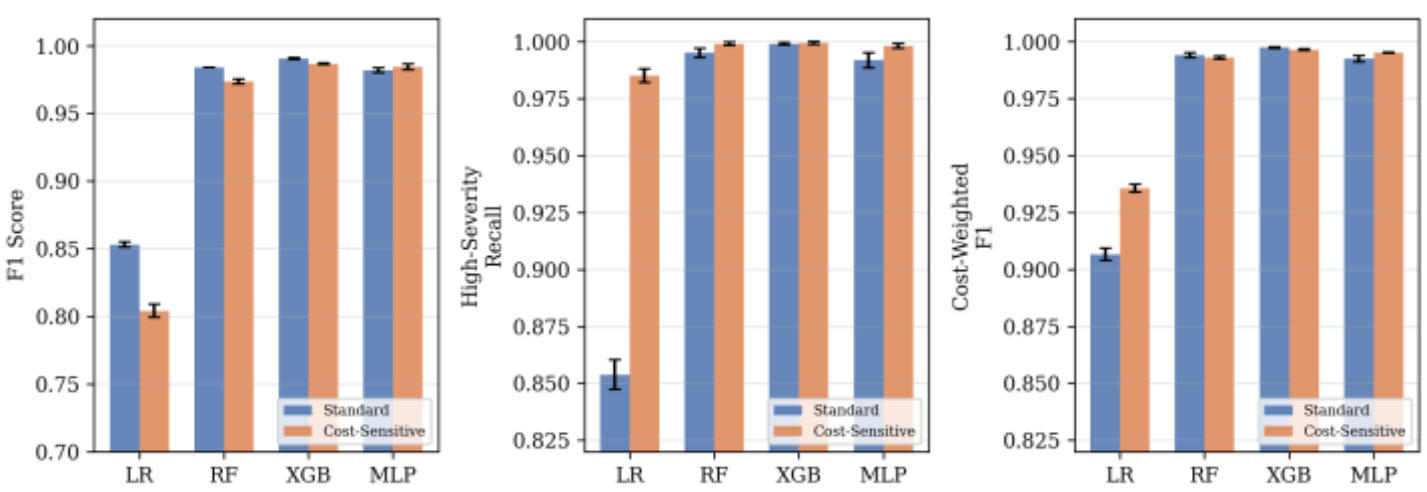

**Fig. 1.** Standard vs. cost-sensitive training. Left: F1. Centre: high-severity recall. Right: cost-weighted F1.

### 5.3 Safety–Efficiency Tradeoff

Cost-sensitive LR shifts sharply upward on the CDR-versus-false-positive plane (CDR 0.860 to 0.977).

### 5.4 Inference Latency

Mean inference time per sample, measured during GroupKFold evaluation, is below 0.001 ms for LR, 0.003 ms for MLP, 0.005 ms for XGBoost and 0.029 ms for Random Forest. All models therefore support real-time operation as a supervisory safety layer.

### 5.5 Sensitivity to λ and Cost Ratios

As λ increases from 0 to 4, HSR rises monotonically from 0.823 to 0.995 whilst F1 decreases from 0.851 to 0.780, confirming λ as an adjustable safety dial. Cost-ratio sensitivity is likewise robust: even the mildest ratio (1:2:5) achieves HSR = 0.960, well above the standard baseline of 0.851, so the conclusion holds regardless of the specific cost values.

### 5.6 Validation on Real Robot Data

We evaluate on the UCI Wall-Following Robot Navigation dataset [23]: 5,456 timestep-level readings from a SCITOS G5 mobile robot following walls clockwise, with 24 circularly arranged ultrasound sensors sampled at 9 Hz. Each timestep carries one of four action labels: Move-Forward, Slight-Right-Turn, Sharp-Right-Turn and Slight-Left-Turn.

The dataset records actions rather than failures, so we map it onto our severity scale: Move-Forward is normal (cost 0); slight turns are routine corrective manoeuvres (cost 1, matching path deviation); Sharp-Right-Turn is emergency obstacle avoidance (cost 5, matching near-miss). No cost-25 level arises as the dataset contains no contact events, so the same 1:5 ratio applies to the two levels present. Sharp-Right-Turn is a successful evasive action, not a failure: the target is the need for emergency intervention, consistent with tracking near-misses as failure precursors.

From sliding windows of 10 timesteps (stride 1) we extract 20 features analogous to the simulation set: range statistics across the sensor ring, sensor variance, rate of change of closest-obstacle distance, directional asymmetries and close-range counts. Each window is labelled with the worst severity in the following 5 timesteps, giving 5,441 samples. As the dataset is a single continuous traversal without episode structure, we use stratified five-fold cross-validation rather than GroupKFold, standardising on the training fold only, with the same $w = 1 + \lambda c$ scheme at $\lambda = 2.0$.

**Table 3.** UCI Wall-Following Robot (SCITOS G5, 24 ultrasound sensors, 5,441 samples).

| Model | Approach | F1 | HSR | CDR | CW-F1 |
|---|---|---|---|---|---|
| LR | Std | 0.787 | 0.679 | 0.686 | 0.808 |
| LR | CS | 0.815 | 1.000 | 1.000 | 0.954 |
| RF | Std | 0.927 | 0.852 | 0.861 | 0.926 |
| RF | CS | 0.918 | 0.990 | 0.991 | 0.978 |
| XGB | Std | 0.955 | 0.929 | 0.932 | 0.963 |
| XGB | CS | 0.949 | 0.989 | 0.989 | 0.984 |
| MLP | Std | 0.917 | 0.885 | 0.891 | 0.937 |
| MLP | CS | 0.917 | 0.885 | 0.891 | 0.937 |

Cost-sensitive LR achieves perfect HSR (1.000) whilst improving F1 (0.787 to 0.815), providing external validation beyond simulation. The MLP is unchanged by cost-sensitive training on this dataset, likely because the limited sample size relative to network capacity means oversampling does not shift the learned boundary.

### 5.7 Adaptive Consequence-Sensitive Ablation

Table 4 reports the five-condition ablation. Moving from class-balanced training through threshold-moving to fixed consequence weighting produces a monotonic increase in high-severity recall (0.851 → 0.918 → 0.985) alongside a monotonic increase in false-positive rate (0.143 → 0.237 → 0.333). Dynamic consequence raises HSR to 0.995 and adding adaptive risk sensitivity reaches 0.998. Missed consequence cost falls from 1,940 to 82, a twenty-fourfold reduction relative to class-balanced training.

**Table 4.** Adaptive-method ablation (Logistic Regression, episode-level GroupKFold).

| Condition | F1 | HSR | CDR | CW-F1 | FPR | Missed |
|---|---|---|---|---|---|---|
| Class-bal. | 0.852 | 0.851 | 0.860 | 0.905 | 0.143 | 1,940 |
| Thresh-move | 0.834 | 0.918 | 0.924 | 0.925 | 0.237 | 1,044 |
| Fixed cons. | 0.804 | 0.985 | 0.977 | 0.935 | 0.333 | 313 |
| Dyn. only | 0.780 | 0.995 | 0.988 | 0.931 | 0.404 | 160 |
| Dyn.+adapt. | 0.748 | 0.998 | 0.994 | 0.921 | 0.499 | 82 |

Two observations require careful interpretation. First, cost-weighted F1 does not increase monotonically: it peaks at fixed consequence weighting (0.935) and declines for dynamic and adaptive variants (0.931 and 0.921). Additional high-severity coverage is purchased at genuine precision cost. Second, much of the raw HSR gain reflects operation at higher false-positive rate rather than improved discrimination.

Table 5 separates these effects using the nested calibration procedure described in Section 4.4. With thresholds selected on held-out calibration episodes and frozen before evaluation, both methods operate at a test FPR of approximately 0.333. Adaptive weighting achieves HSR = 0.986 ± 0.002 against 0.984 ± 0.003 for fixed weighting, with missed cost 306 versus 318 and identical CW-F1 (0.935). The discriminative advantage is therefore real but small; the substantial missed-cost reduction in Table 4 is primarily attributable to the adaptive method selecting a more conservative operating point rather than to better ranking of samples.

**Table 5.** Matched-FPR comparison with nested threshold calibration (test FPR ≈ 0.333).

| Method | HSR | CW-F1 | Missed | Test FPR |
|---|---|---|---|---|
| Fixed cons. | 0.984 | 0.935 | 318 | 0.333 |
| Dyn.+adapt. | 0.986 | 0.935 | 306 | 0.334 |

The five conditions trace a safety coverage frontier: each successive method extends high-severity recall at the cost of additional false alarms, with a corresponding reduction in undetected consequence cost. Where collisions carry catastrophic cost and false alarms are merely disruptive, operating further along this frontier is rational.

### 5.8 Paired Fold-Level Comparison

Aggregate means can conceal inconsistency. Table 6 reports paired differences between adaptive and fixed weighting on each held-out fold. Adaptive weighting achieves higher HSR, higher CDR and lower missed cost in all five folds, with ΔHSR ranging from +0.011 to +0.016 and missed cost reduced by 175 to 264 units per fold. The direction of effect is therefore consistent rather than driven by a single favourable partition.

**Table 6.** Paired fold-level differences (adaptive − fixed), episode-level GroupKFold.

| Fold | ΔHSR | ΔCDR | ΔMissed | ΔFPR |
|---|---|---|---|---|
| 0 | +0.015 | +0.019 | −252 | +0.179 |
| 1 | +0.012 | +0.018 | −212 | +0.162 |
| 2 | +0.016 | +0.017 | −264 | +0.161 |
| 3 | +0.011 | +0.013 | −175 | +0.174 |
| 4 | +0.014 | +0.018 | −248 | +0.150 |

### 5.9 Coefficient Robustness

The context coefficients were fixed a priori. To confirm results do not depend on that setting, we scaled the entire coefficient vector by 0.5× to 1.5× and re-ran the full evaluation. Table 7 shows HSR varies only between 0.995 and 0.999 across this threefold span. Larger coefficients produce monotonically more conservative behaviour, as the formulation predicts, with no anomalous configurations. Coefficients were not selected on the basis of these results.

**Table 7.** Coefficient robustness (0.75× and 1.25× omitted; both interpolate monotonically).

| Scale | HSR | CDR | Missed | FPR | F1 |
|---|---|---|---|---|---|
| 0.50× | 0.995 | 0.989 | 142 | 0.427 | 0.772 |
| 1.00× | 0.998 | 0.994 | 82 | 0.499 | 0.748 |
| 1.50× | 0.999 | 0.995 | 67 | 0.548 | 0.732 |

### 5.10 Risk-Stratified Analysis

The central claim of the adaptive formulation is that conservatism should be allocated according to physical operating state. To test this mechanism directly rather than only its aggregate effect, we partitioned test samples into terciles by each contextual variable (proximity risk, velocity and sensing uncertainty), using thresholds computed from the training distribution, and compared fixed against adaptive weighting within each band.

The result (Table 8) is partly counter-intuitive. For proximity risk and sensing uncertainty, the largest adaptive gains occur in the low-risk bands: HSR rises from 0.768 to 0.973 (+0.205) at low proximity risk and from 0.861 to 0.984 (+0.122) at low sensing uncertainty, whilst in the corresponding high-risk bands both methods reach HSR $\geq$ 0.998 and the difference is negligible. Fixed weighting saturates where danger is obvious; the stratified analysis suggests adaptive training improves detection of high-severity events even when individual contextual indicators are not themselves extreme. Velocity matches the original intuition, with the high-velocity band showing the largest gain (+0.065 against +0.003 at low velocity).

**Table 8.** Risk-stratified comparison (low and high terciles). Δ = adaptive − fixed.

| Variable | Band | HSR fix | HSR adapt | ΔHSR | ΔMissed |
|---|---|---|---|---|---|
| Proximity | Low | 0.768 | 0.973 | +0.205 | −152 |
| Proximity | High | 0.998 | 1.000 | +0.002 | −36 |
| Velocity | Low | 0.993 | 0.996 | +0.003 | −57 |
| Velocity | High | 0.933 | 0.998 | +0.065 | −111 |
| Sensing | Low | 0.861 | 0.984 | +0.122 | −158 |
| Sensing | High | 0.999 | 1.000 | +0.001 | −19 |

### 5.11 Collision-Inclusive Evaluation

The primary simulation produced no collisions, so the cost-25 severity level was not empirically exercised. We therefore constructed a controlled extension using a bounded repulsive term, allowing the attractive term to overwhelm it in cluttered scenes. Episodes mix a nominal configuration (75%) with a degraded one having weakened repulsion, more obstacles and higher speed (25%). All severity thresholds, features, costs and the GroupKFold protocol are identical to the primary study; only the repulsion model changes.

This yields 600 episodes (188,728 timesteps): 92.53% normal, 3.48% path deviation, 3.93% near-miss and 0.06% collision, with 108 episodes terminating in contact and 1,341 cost-25 samples among 60,114 windows (14.8% failure rate). Table 9 reports collision recall, recall restricted to cost-25 samples, and the missed cost attributable specifically to collisions.

**Table 9.** Collision-inclusive evaluation. Collision recall is restricted to cost-25 samples.

| Method | HSR | Coll. rec. | CDR | Missed | Missed coll. | FPR |
|---|---|---|---|---|---|---|
| Class-balanced | 0.667 | 0.850 | 0.765 | 2,745 | 1,000 | 0.235 |
| Fixed cons. | 0.821 | 0.966 | 0.888 | 1,312 | 225 | 0.413 |
| Dyn.+adapt. | 0.978 | 0.984 | 0.982 | 215 | 105 | 0.799 |

Collision recall increases monotonically (0.850 → 0.966 → 0.984) with missed collision cost falling 1,000 → 225 → 105. The highest severity level therefore benefits in the same direction as the near-miss results, which the primary simulation could not demonstrate.

The false-positive cost must be stated plainly: adaptive weighting operates at FPR = 0.799 on this harder task against 0.413 for fixed weighting, flagging most windows as risky. Such an operating point is defensible only where collisions are genuinely catastrophic relative to interruption; for most deployments fixed weighting (FPR 0.413, collision recall 0.966) is more practical. The adaptive formulation's value is making this trade explicit and controllable.

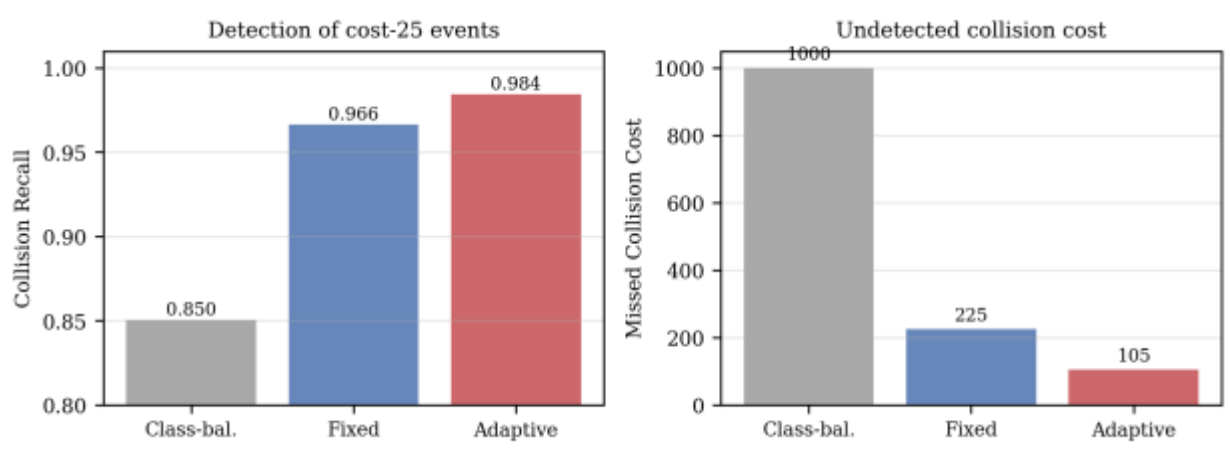

**Fig. 2.** Collision-specific detection on cost-25 samples. Left: recall rises 0.850 → 0.966 (fixed) → 0.984 (adaptive), the last at FPR 0.799. Right: undetected collision cost falls 1,000 → 105.

## 6 Discussion

### 6.1 The Case for Cost-Sensitive Evaluation

Our results reveal a significant disconnect between standard and cost-sensitive evaluation. Under standard metrics, Logistic Regression appears to be a reasonable performer (F1 = 0.853, AUROC = 0.943). However, cost-sensitive analysis reveals that it misses 14.9% of high-severity failures, events with five times the consequence cost of path deviations, amounting to 1,940 units of missed consequence cost. A deployment decision based solely on F1 would not capture this safety gap; this echoes the healthcare AI literature [3], where aggregate metrics conceal allocation-relevant failures.

### 6.2 Architecture-Dependent Sensitivity

The observation that cost-sensitive training has the largest effect on Logistic Regression and the smallest on XGBoost is itself a useful finding. XGBoost's inherent capacity for learning complex decision boundaries, combined with its native handling of class imbalance through positive-class scaling, means it already performs near-perfectly on high-severity events in standard mode. Logistic Regression, by contrast, has a linear decision boundary that must balance competing demands. Cost-sensitive sample weighting reshapes the effective loss landscape, shifting the decision boundary to favour the detection of high-consequence events at the cost of increased false positives on lower-severity events.

### 6.3 What Adaptive Weighting Does and Does Not Provide

The ablation supports a nuanced conclusion. Context-dependent consequence modelling provides additional high-severity coverage at more conservative operating points: at its natural operating point, adaptive weighting reduces missed consequence cost from 313 to 82, a fourfold reduction. For applications where undetected high-severity events carry catastrophic cost, this is meaningful.

However, the matched-FPR analysis shows most of this improvement stems from operating at a more conservative threshold rather than superior discrimination. At equal false-alarm rates, with thresholds calibrated on held-out episodes, the advantage narrows to 0.986 versus 0.984 HSR and cost-weighted F1 is identical (0.935). The adaptive formulation therefore provides a principled mechanism for selecting where on the safety frontier to operate, becoming more conservative when velocity is high, obstacles are close and sensing is degraded, rather than a fundamentally better classifier.

The risk-stratified analysis (Section 5.10) clarifies where this matters. Fixed weighting already saturates in states that are obviously hazardous by proximity or sensing quality, leaving no headroom; the adaptive method's contribution is concentrated in states that appear benign on those axes but nonetheless precede high-severity events. Velocity is the exception, showing the expected pattern of larger gains at higher speed. We note that the adaptive weighting uses these same contextual variables, so this analysis characterises the model's behaviour rather than establishing a causal account of why those events occur; trajectory-level analysis would be required for the latter.

### 6.4 Implications for Robotic Deployment

The practical implication of our findings is that a cost-sensitive failure prediction module could serve as a safety layer for autonomous navigation systems. Operating alongside any existing navigation stack, such a module would monitor incoming sensor and motion data and raise alerts when the predicted risk of a high-consequence failure exceeds a threshold. The cost-sensitive training ensures that this module prioritises the detection of events that matter most, near-misses and collisions, rather than optimising for overall failure prediction accuracy.

The $\lambda$ analysis provides an adjustable safety dial, and the dominant features, current minimum range (importance 0.412) and sensor noise level (0.262), are available in real-time on most platforms with sub-millisecond inference. The prominence of sensor noise level suggests self-aware sensing is an underappreciated component of safe navigation.

## 7 Limitations and Future Work

The primary evaluation is in simulation; although externally validated on the UCI SCITOS G5 dataset, prospective deployment on a physical platform remains necessary. The three-level taxonomy could extend to continuous cost functions. The handcrafted feature set is partly tied to the potential-field controller, requiring adaptation to latent representations for learning-based controllers. Context parameters were fixed a priori rather than optimised. The collision-inclusive results come from a modified controller rather than the primary simulation, so the two sets of numbers characterise different operating regimes and should not be compared directly.

Future work should address validation on physical platforms using datasets such as TUM RGB-D [21] or EuRoC MAV [22], integration with reinforcement learning controllers for closed-loop risk-aware navigation, and extension to multi-robot systems where failure cost depends on spatial configuration.

## 8 Conclusion

This paper has reframed navigation failure prediction as consequence-sensitive forecasting. Fixed consequence weighting raises Logistic Regression high-severity recall from 0.851 to 0.985 whilst reducing missed consequence cost from 1,940 to 313. The adaptive extension, which scales consequence by velocity, proximity and sensing uncertainty, reaches high-severity recall of 0.998 with missed cost of 82. The UCI SCITOS G5 dataset provides external validation, with cost-sensitive LR reaching perfect high-severity recall.

These gains carry a measurable cost. Cost-weighted F1 peaks at fixed weighting (0.935) rather than adaptive (0.921) and under matched false-positive conditions the advantage narrows to 0.986 versus 0.984. The adaptive formulation is best understood as a mechanism for allocating conservatism by physical context rather than a uniformly superior classifier. The effect is consistent (5/5 folds) and stable across a threefold coefficient span. A collision-inclusive extension confirms the highest severity level benefits in the same direction, collision recall 0.984, missed collision cost falling from 1,000 to 105, at a false-positive rate defensible only where contact is catastrophic.

We argue that the robotics community should adopt safety-oriented evaluation metrics alongside standard ones. Not all errors are equal and our evaluation frameworks should reflect this.

**Disclosure of Interests.** The author has no competing interests to declare that are relevant to the content of this article.